\pdfoutput=1

\documentclass[11pt]{article}

\usepackage[preprint]{acl}

\usepackage{times}
\usepackage{latexsym}

\usepackage[T1]{fontenc}

\usepackage[utf8]{inputenc}

\usepackage{microtype}

\usepackage{inconsolata}

\usepackage{graphicx}
\usepackage{booktabs}
\usepackage{colortbl}

\usepackage[most]{tcolorbox}
\usepackage{float}
\usepackage{xspace}
\tcbset{
aibox/.style={
width=430pt,
top=10pt,
colback=white,
colframe=black,
colbacktitle=black,
enhanced,
center,
attach boxed title to top left={yshift=-0.12in,xshift=0.15in},
boxed title style={boxrule=0pt,colframe=white},
}
}
\newtcolorbox{AIbox}[2][]{aibox,title=#2,#1}

\title{Reasoning on Multiple Needles In A Haystack}

\author{Yidong Wang \\
Independent Researcher \\
  \texttt{yidongw2005@163.com}}

\begin{document}
\maketitle
\begin{abstract}
The Needle In A Haystack (NIAH) task has been widely used to evaluate the long-context question-answering capabilities of Large Language Models (LLMs). However, its reliance on simple retrieval limits its effectiveness. To address this limitation, recent studies have introduced the \textit{Multiple Needles In A Haystack Reasoning} (MNIAH-R) task, which incorporates supporting documents (\textit{Multiple needles}) of multi-hop reasoning tasks into a distracting context (\textit{Haystack}). Despite this advancement, existing approaches still fail to address the issue of models providing direct answers from internal knowledge, and they do not explain or mitigate the decline in accuracy as context length increases. In this paper, we tackle the memory-based answering problem by filtering out direct-answer questions, and we reveal that performance degradation is primarily driven by the reduction in the length of the thinking process as the input length increases. Building on this insight, we decompose the thinking process into retrieval and reasoning stages and introduce a reflection mechanism for multi-round extension. We also train a model using the generated iterative thinking process, which helps mitigate the performance degradation. Furthermore, we demonstrate the application of this retrieval-reflection capability in mathematical reasoning scenarios, improving GPT-4o's performance on AIME~2024.

\end{abstract}

\section{Introduction}

With advancements in context window extension technologies~\citep{dao2022flashattention, chen2023extending, xiong2023effective, bai2024longalign}, models such as Qwen2.5~\citep{qwen_2p5_1m} now support context windows of up to 1M tokens. However, the Needle In A Haystack~(NIAH) task~\citep{NIAH}, once a key benchmark for long-context evaluation, has become less effective~\citep{vodrahalli2024michelangelo, hsieh2024ruler}. While new benchmarks~\citep{zhang2024inftybenchextendinglongcontext, NOCHA, bai2023longbench} aim to assess long-context understanding, their scalability issues limit their ability to evaluate models with very large context windows.

Recent works~\citep{hsieh2024ruler, Needlebench, vodrahalli2024michelangelo} introduced the Multiple Needles In A Haystack Reasoning~(MNIAH-R) task as a scalable diagnostic task for reasoning, which incorporates supporting documents of multi-hop reasoning tasks within a pool of distracting information. However, these studies did not provide a comprehensive analysis of the underlying causes of the accuracy decline with increasing context length, nor did they explore potential solutions to address this issue. Additionally, the problem of models relying on internal knowledge for direct answers was not sufficiently addressed.

To address the memory-based answering issue, we evaluate models by focusing on questions where models answer correctly based on supporting documents in multi-hop reasoning tasks, but incorrectly when answering directly, ensuring that models do not rely on internal knowledge. The experimental results show that, before filtering, models' accuracy decreases slightly with fluctuations as the context length increases, with minimal differences in the rate of decline between models. However, after filtering, the accuracy drops significantly as the context length increases, with open-source models experiencing a greater decline than commercial models.

We investigate the causes of accuracy decline and find that it is not related to needles placement or the distance between them, but rather to the reduction in thinking process length as input length increases. Based on these observations, we decompose the thinking process into retrieval and reasoning stages and introduce a reflection mechanism for multi-round extension, exploring the Test-Time Scaling Law~\citep{snell2024scalingllmtesttimecompute}. On this basis, we train a model using the generated iterative thinking process, reducing the accuracy drop from 25.8\% to 4.6\%. Additionally, we apply the retrieval-reflection capability to a mathematical reasoning context, improving the Pass@1 score of GPT-4o on AIME~2024 from 9.3 to 15.3.

Our contributions are as follows: 
\begin{itemize}
    \item We demonstrate that memory-based responses significantly impact MNIAH-R performance. After filtering, we assess models' capabilities, highlighting a performance gap between open-source and commercial models.
    \item We identify that the accuracy drop is related to a shorter thinking process, which can be mitigated by decoupling retrieval from inference and incorporating a reflection mechanism for iterative thinking.
    \item We train a model with retrieval-reflection ability and apply it to mathematical scenarios.
\end{itemize}


\section{MNIAH-R}

\textbf{Dataset Setup} For multi-hop reasoning tasks, we use the HotpotQA~\citep{HotPotQA} and IRE~\citep{IRE} datasets. HotpotQA contains 113k wikipedia-based question-answer pairs requiring reasoning over multiple documents, and we focus on its \texttt{dev\_distractor} subset, which includes clearly defined standard answers and their corresponding supporting documents. The IRE dataset introduces a stepwise counterfactual benchmark with 782 factual and counterfactual instances to assess multi-step reasoning. Please refer to Appendix~\ref{sec:cons_MNIAH-R} for more details about dataset.

\noindent \textbf{Model Evaluation} We select six long-context models: GPT-4o~\citep{GPT-4o}, GPT-4o-mini~\citep{GPT-4o-mini}, Claude-3.5-Sonnet~\citep{Claude_3.5_Sonnet}, Llama-3-8B-ProLong-64k-Instruct~\citep{Llama-3-8B-ProLong-64k-Instruct}, Qwen-2.5-1M~\citep{qwen_2p5_1m}, and GLM-4-9B-Chat-1M~\citep{glm-4-9b-chat-1m}. We use greedy decoding for inference and employ DeepSeek-V3~\citep{guo2025deepseek} to assess the correctness of model responses. For further details on the models introduction and evaluation prompt, please refer to Appendix~\ref{sec:eval_MNIAH-R}.

\noindent \textbf{Dataset Filtering} To tackle the issue of memory-based answering, we evaluate models by concentrating on questions where models can answer correctly when supported by the provided documents, but fail to do so when answering directly without relying on external context. By emphasizing these types of questions, we mitigate the tendency of the model to recall answers from previous training data and provide direct responses. For a detailed breakdown of the data statistics before and after filtering, please refer to Appendix~\ref{sec:appendix_Statistics on Filtered questions}.

\begin{figure}
    \centering
    \includegraphics[width=1\linewidth]{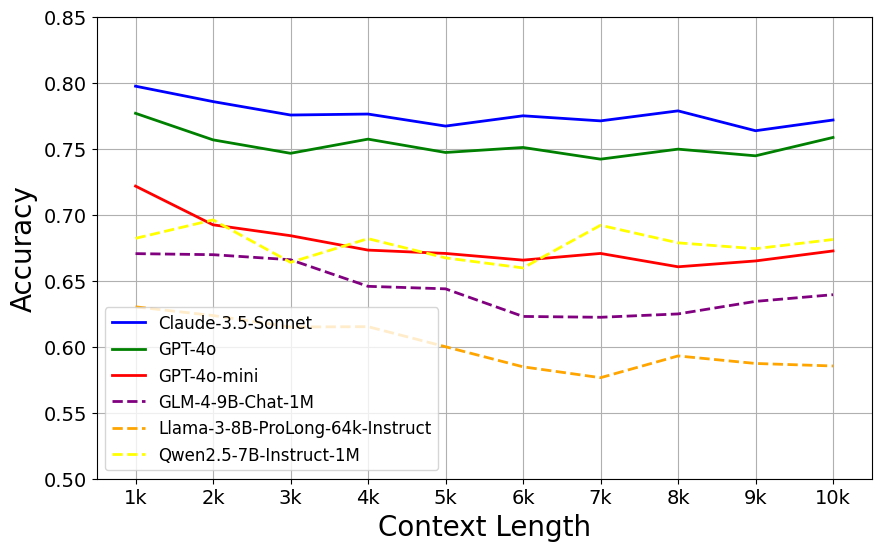}
    \caption{Performance on MNIAH-R \textit{before filtering}.}
    \vspace{-2mm}
    \label{fig:Universal}
\end{figure}
\begin{figure}
    \centering
    \includegraphics[width=1\linewidth]{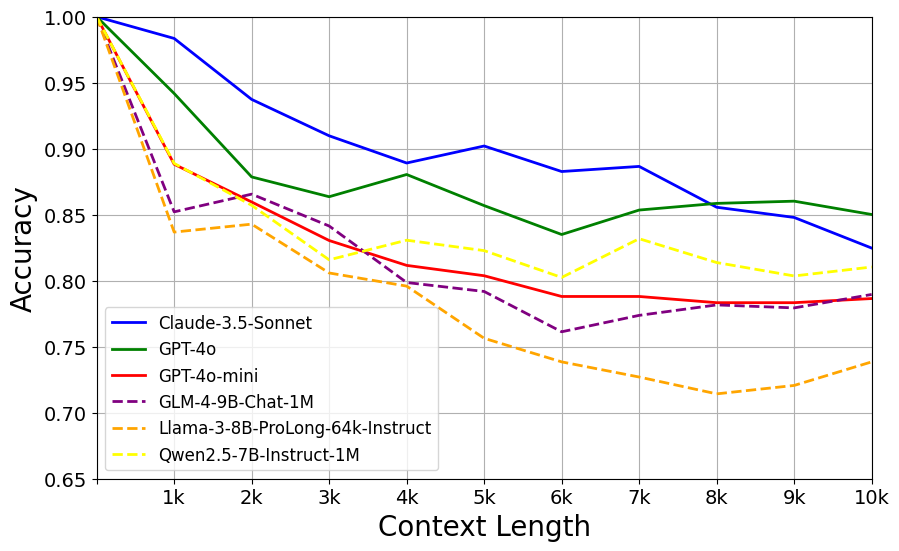}
    \caption{Performance on MNIAH-R \textit{after filtering}.}
    \vspace{-2mm}
    \label{fig:MNIAH_R_after_filtering}
\end{figure}

\noindent \textbf{Results Analysis} As shown in Figure~\ref{fig:Universal} and \ref{fig:MNIAH_R_after_filtering}, before filtering, although there is a noticeable difference in accuracy between individual models, the decrease in accuracy with increasing context length is negligible. This does not adequately reflect the impact of context length on the performance of different models. After filtering, we observe a clear decline in accuracy for each model as context length increases. Notably, the open-source model, represented by the dashed line, exhibits a steeper decline compared to the closed-source model, represented by the solid line.

\section{Explanations for Accuracy Decrease}


To explain the decrease in accuracy with increasing context length, we further investigate the intersection of the models' filtered questions. First, we find that the accuracy at the intersection follows a similar decreasing trend with context length; please see the Appendix~\ref{sec:appendix_Statistics on Filtered questions} for details. Building on this, we explore three potential factors inspired by related works: \textit{needles placement}~\citep{LostInTheMiddle}, \textit{distance between Needles}~\citep{Multi_Needle}, and \textit{thinking process length}~\citep{team2025kimi}.

\begin{figure}
    \centering
    \includegraphics[width=1\linewidth]{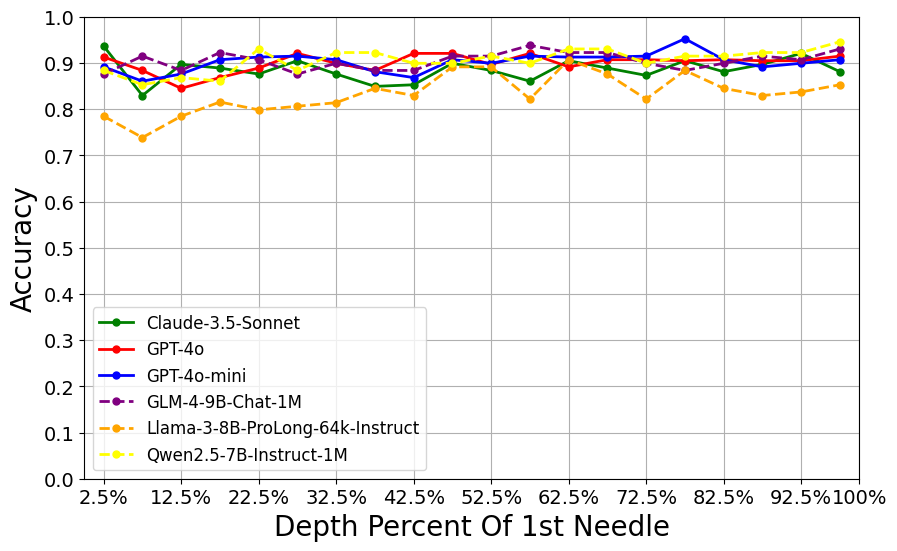}
    \caption{Impact of \textit{Needles Placement Positions}.}
    \label{fig:Needles_replacement}
\end{figure}

\noindent \textbf{Needles Placement} We first examine the impact of supporting documents (\textit{needles}) placement within the context on accuracy decline. Needle position is measured using \textit{depth percent}, which indicates the relative position of a needle within the context window as a percentage, with 0\% representing the start and 100\% the end. Each question of dataset contains two needles placed at fixed intervals of 500 tokens. The position of the first needle varies from 2.5\% to 97.5\% in 10\% increments within the 10k-token context window, and accuracy is compared accordingly. The experimental results, shown in Figure~\ref{fig:Needles_replacement}, reveal that accuracy fluctuates slightly with changes in needle placement and exhibits no clear trend, suggesting that accuracy degradation is independent of needle placement.

\begin{figure}
    \centering
    \includegraphics[width=1\linewidth]{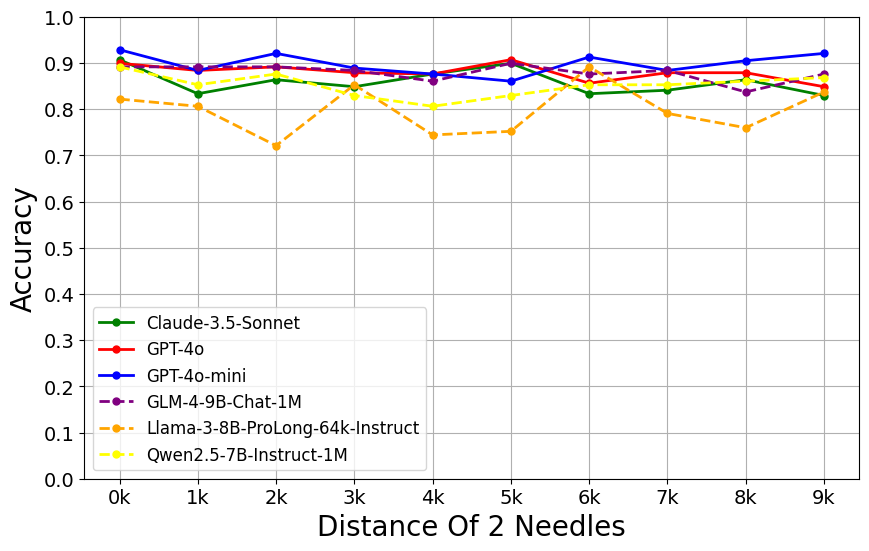}
    \caption{Impact of \textit{Distance Between Needles}.}
    \label{fig:Distance_between_needles}
\end{figure}

\noindent \textbf{Distance Between Needles} We further investigate the impact of the distance between needles on accuracy decline. After confirming that needle placement has no effect, we fix the first needle at the 250-token position within a 10k-token context. We then vary the position of the second needle, increasing the distance between the two needles by 1k tokens at each step, up to 9k tokens, and compare the accuracy. As shown in Figure~\ref{fig:Distance_between_needles}, no significant accuracy trend is observed with increasing distance, suggesting that the accuracy decline is not related to the distance between the needles.

\begin{figure}[!h]
    \centering
    \includegraphics[width=1\linewidth]{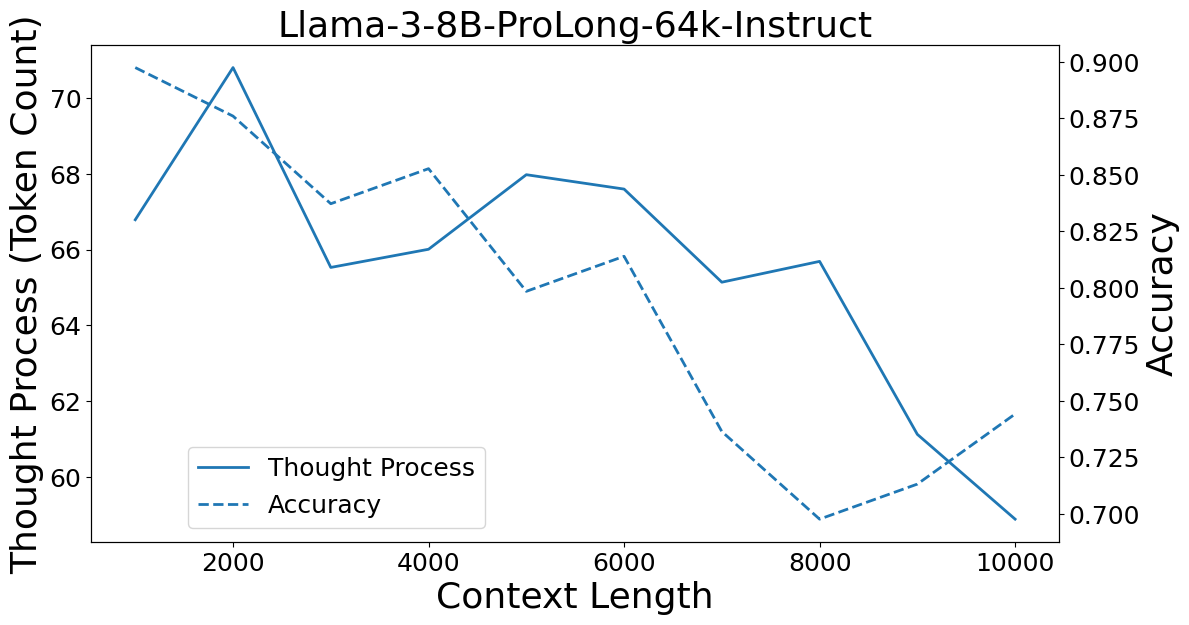}
    \caption{Impact of context length on \textit{thinking process length} for model with significant accuracy decline.}
    \label{fig:Llama_prolong_thinking_length}
\end{figure}

\begin{figure}[!h]
    \centering
    \includegraphics[width=1\linewidth]{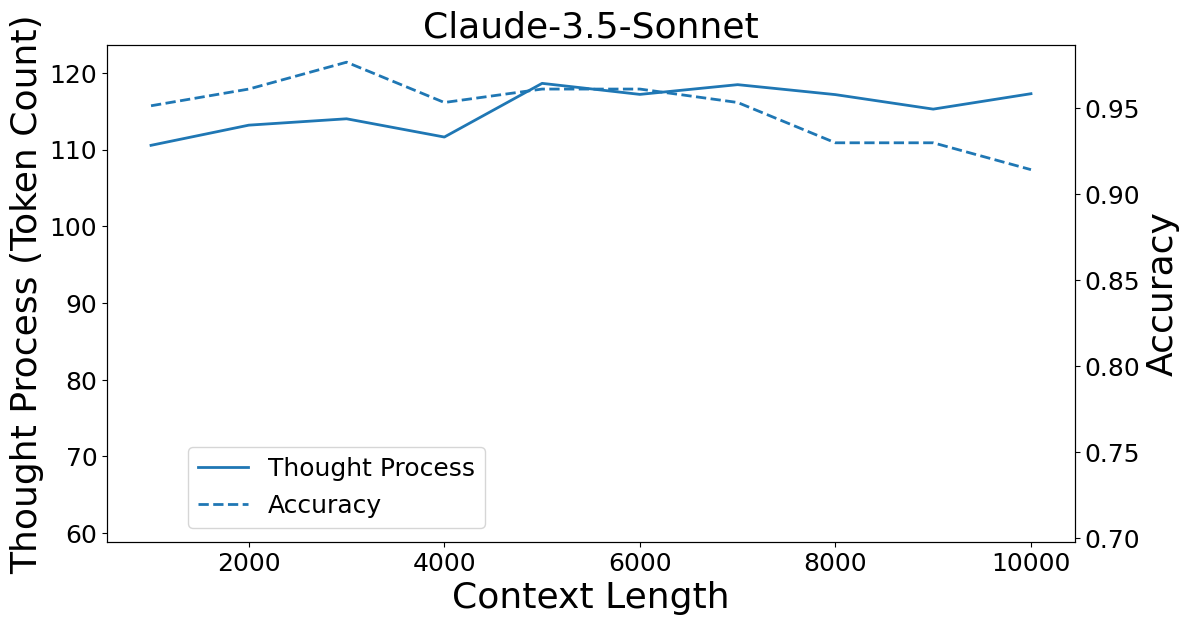}
    \caption{Impact of context length on \textit{thinking process length} for model with minimal accuracy decline.}
    \label{fig:claude_thiking_length}
\end{figure}

\noindent \textbf{Thinking Process Length} We continue investigate how increasing context length affects the thinking process. We instruct models to provide their thinking processes before answering and then count the number of thinking processes' tokens. The prompt is shown in Appendix~\ref{sec:appendix_Prompts}. Two models are examined: \texttt{Claude-3.5-Sonnet}, which has the smallest accuracy decline, and \texttt{Llama-3-8B-ProLong-64k-Instruct}, which has the largest. As shown in Figures~\ref{fig:Llama_prolong_thinking_length} and \ref{fig:claude_thiking_length}, the length of model's thinking process is strongly correlated with its response accuracy, consistent with recent findings on mathematical tasks~\citep{team2025kimi}. The accuracy decline with increasing context length may be due to the reason that longer contexts shorten the thinking process, leading to incomplete or incorrect information retrieval.

\section{Decrease Mitigation and Application}

\begin{figure*}[!h]
    \centering
    \includegraphics[width=1\linewidth]{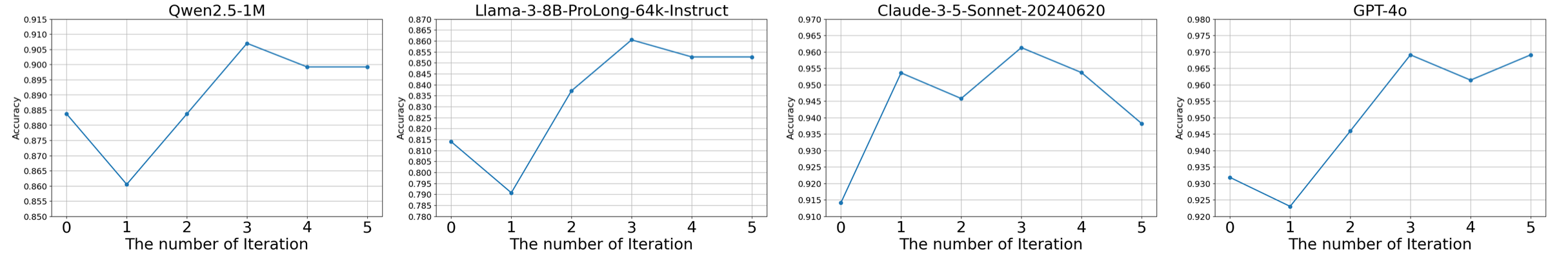}
    \caption{Mitigation of the accuracy decline with increased iterations of thinking process.}
    \label{fig:10R}
\end{figure*}

\label{sec:Multi-Round Retrieval and Reasoning}

\noindent \textbf{Test-Time Scaling} To address performance degradation, we propose a strategy that extends the model's reasoning process by dividing it into two stages: information retrieval and reasoning. After reasoning, a reflection phase re-evaluates and supplements the retrieved information through iterative steps. We perform five iterations, as shown in Figure~\ref{fig:10R}, with accuracy improving and the rate of decline decreasing with each iteration. However, after the third iteration, performance on the MNIAH-R task plateaus, indicating a saturation point in scaling. Detailed generation settings can be found in Appendix~\ref{sec:appendix_Generation Settings of All Experiments}, and the prompts are outlined in Appendix~\ref{sec:appendix_Prompts}.

\noindent \textbf{Training with Iterative Thinking Process} Since model performance typically plateaus after the third iteration, we select the first two rounds of iterative thinking Process from GPT-4o, which show the smallest accuracy drop on the MNIAH-R task, to construct the fine-tuning dataset. The filtered questions from GPT-4o consist of 594 items, from which we randomly sample 416 for the \texttt{\textbf{Training}} set and 178 for the \texttt{\textbf{Test}} set, as detailed in Appendix~\ref{sec:appendix_Construction_details_of_fine-tuning_dataset} and \ref{sec:appendix_Training Setting Details}. We then fine-tune the \texttt{Llama-3-8B-ProLong-64k-Instruct} model, which shows the greatest accuracy decline, and evaluate its performance on the test set. The results, shown in Table~\ref{tab:mitigation}, indicate that fine-tuning reduces the accuracy drop from 25.8\% to 4.6\%, significantly outperforming direct fine-tuning with original answers.

\begin{table}[]
 \addtolength\tabcolsep{2.0pt}
\begin{tabular}{lrrr}
\toprule
Models    & 1K   & 10K  & \textbf{$\Delta$} \\ \midrule
Llama-3-8B-ProLong   & 85.7 & 59.9 &
25.8    \\ \midrule
\rowcolor{gray!15}\multicolumn{4}{c}{\textbf{Trained}} \\
w/ Direct Answer& 87.2 & 63.3 & 23.9   \\ 
w/ Thinking Process & \textbf{89.3} & \textbf{84.7} & \textbf{4.6}     \\ 
\bottomrule
\end{tabular}
\vspace{-2mm}
\caption{Comparison of accuracy on \texttt{\textbf{Test}} set. \textbf{$\Delta$} denotes the decrease from 1k to 10k. \textit{w/ Direct Answer} refers to train with the original answer of multi-hop questions, while \textit{w/ Thinking Process} represents training with responses extracted from iterative thinking process.}
\vspace{-1mm}
\label{tab:mitigation}
\end{table}

\begin{table}[]
 \addtolength\tabcolsep{20.0pt}
\begin{tabular}{lr}
\toprule
Models &  \begin{tabular}[c]{@{}r@{}}AIME2024\\ (pass@1)\end{tabular}\\ \midrule
GPT-4o   & 9.3\\ \midrule
\rowcolor{gray!15}\multicolumn{2}{c}{\textbf{Extracted by Llama-3-8B-prolong}} \\
w/o Training & 10.0\\
w/ Training & \textbf{15.3}\\
\bottomrule
\end{tabular}
\vspace{-1mm}
\caption{Applying models to mathematical scenarios by extracting correct solution. \textit{w/ Train} refers to the model fine-tuned with iterative thinking process.}

\label{tab:math}
\end{table}


\noindent \textbf{Mathematical Application} Building on the improvements in MNIAH-R task, we further apply this retrieval-reflection capability to mathematical scenarios. We first test GPT-4o on AIME~2024, requiring it to provide detailed intermediate steps, and generating five responses per question. The five solutions are then combined with the original problem and fed into the trained model to extract the correct answers. Given that many of the solutions are incorrect, this provides an opportunity to demonstrate the practical application of our model's retrieval and reasoning capabilities. To mitigate output repetition~\citep{guo2025deepseek}, we adjust the \texttt{temperature} to 0.6 and \texttt{top\_p} to 0.95, and also generating five responses per query. The evaluation metric is pass@1~\citep{chen2021evaluating}. As shown in Table~\ref{tab:math}, the fine-tuned model boosts GPT-4o’s pass@1 score from 9.3 to 15.3 by correctly extracting the right solution, outperforming the original model's score of 10.0, demonstrating the effectiveness of our trained model in mathematical applications.


\section{Conclusion}

We conduct a thorough study of the MNIAH-R task, addressing models' reliance on internal knowledge instead of context through question filtering, and identify a performance gap between open-source and commercial models. We find that accuracy declines due to a shortened thinking process, rather than the placement or distance of supporting documents. We decompose the thinking process into retrieval and reasoning stages, incorporating reflection in multi-round iterations to mitigate this decline. Fine-tuning the model with iterative thinking steps significantly mitigates the decline and demonstrates application in mathematical scenarios.

\clearpage
\section*{Limitations}
While we investigate the significant performance degradation of certain long-context models on the MNIAH-R task, there are still novel reasoning models that remain underexplored. In mathematical reasoning scenarios, we observe that while the trained models improve the performance of GPT-4o, their impact on more powerful models like o3-mini is less pronounced. We hypothesize that this may be due to the weaker reasoning capabilities of these models prior to training, as well as the inherent difficulty of the AIME questions. Future research may focus on training more advanced reasoning models and further exploring the potential benefits of retrieval-reflection approaches on mathematical reasoning tasks across a broader range of problems.

\bibliography{custom}

\newpage
\appendix
\label{sec:appendix}

\section{Related Work}

\paragraph{Long-context Language Models}

Recent advancements in long-context language models have focused on reducing memory requirements and improving scalability. Techniques such as Flash attention~\citep{fa,fa2} and Ring attention~\citep{ra} are designed to handle longer contexts with lower memory consumption. Sparse attention methods~\citep{longlora, longnet} and new position embedding techniques~\citep{alibi,xpos,rope} further enhance context processing. In addition, methods like recurrence mechanisms for context caching~\citep{beacon,rmt} and retrieval-based strategies~\citep{longmem, infllm} aim to improve efficiency. Alternative architectures such as Mamba~\citep{mamba} and RWKV~\citep{rwkv} also offer efficient handling of extended contexts.

\paragraph{Benchmarks for Long-context Evaluation}

Several benchmarks evaluate long-context models, focusing on tasks like retrieval, summarization, and reasoning. ZeroSCROLLS~\citep{zeroscrolls} includes realistic tasks such as long-document QA and summarization, while LongBench~\citep{bai2023longbench} and InfiniteBench~\citep{zhang2024inftybenchextendinglongcontext} support tasks with contexts exceeding 100K tokens. Synthetic benchmarks offer more flexibility in task design, enabling analysis of scaling behavior and model capabilities in long-range discourse modeling~\citep{chapterbreak} and in-context learning~\citep{manyiclgdm}. Additionally, the MNIAH-R task is listed as an evaluation task in several studies like Ruler~\citep{hsieh2024ruler} and Michelangelo~\citep{vodrahalli2024michelangelo}, which conclude that the model's accuracy on this task decreases as context length increases. However, these studies does not explain this phenomenon or propose any mitigation strategies.

\paragraph{Challenges in Long-context Reasoning Evaluation}

Despite progress, evaluating long-context reasoning remains challenging. Many existing benchmarks suffer from issues like "short-circuiting"~\citep{vodrahalli2024michelangelo}, where models can bypass the need for full context~\citep{kuratov2024babilongtestinglimitsllms}, and "secret retrieval tasks"~\citep{vodrahalli2024michelangelo}, where models perform well by retrieving information rather than synthesizing it~\citep{hsieh2024ruler}. Additionally, out-of-distribution distractors~\citep{Needlebench} can simplify tasks by making relevant information easily identifiable. For the "short-circuiting" issue, although it can be mitigated to some extent by using counterfactual datasets~\citep{IRE}, there are still some problems due to limited data quality and potential data leakage.

\section{Details on MNIAH-R Task}
\label{sec:appendix_MNIAH-R}

\subsection{Construction of MNIAH-R Task}
\label{sec:cons_MNIAH-R}
In the MNIAH-R task, the questions are sourced from two sets: 782 questions from the IRE~\citep{IRE} dataset and 800 randomly sampled questions from the \texttt{dev\_distractor} subset of HotpotQA~\citep{HotPotQA}, with the latter matching the size of the IRE dataset. Each question in both datasets is paired with two supporting documents and eight distractor paragraphs, which are related but do not contain any supporting facts. For haystack creation and needle insertion, we follow the approach outlined in previous work~\citep{NIAH}, using the PaulGrahamEssays dataset as the haystack to extend the input to the target length. For needle insertion, we randomly and evenly insert the aforementioned 10 passages into the haystack for evaluation.

\subsection{Model Evaluation Details}
\label{sec:eval_MNIAH-R}
For detailed descriptions of models, GPT-4o~\citep{GPT-4o} is a multimodal AI model for natural interaction, excelling in text, audio, and image processing with a context window of 128 tokens; 

GPT-4o-mini~\citep{GPT-4o-mini} is a cost-efficient, lightweight AI model capable of processing a 128k tokens context length; 

Claude 3.5 Sonnet~\citep{Claude_3.5_Sonnet} is a high-performance AI model by Anthropic, excelling in reasoning and coding with 200k-token context processing and enhanced speed; 

Llama-3-8B-ProLong-64k-Instruct~\citep{Llama-3-8B-ProLong-64k-Instruct} is a long-context optimized model with state-of-the-art performance on 64k context tasks; 

Qwen2.5-7B-Instruct-1M~\citep{qwen_2p5_1m} is a long-context model with 1 million token capacity, optimized for superior performance in extended context tasks; 

GLM-4-9B-Chat-1M~\citep{glm-4-9b-chat-1m} is an advanced open-source model supporting 1 million token context length, excelling in long-text reasoning and multilingual dialogue. 

For evaluating the correctness of models' answers using DeepSeek-V3~\citep{guo2025deepseek}, we combine the problem and the standard answer to ensure a more accurate judgment. The prompt is illustrated in Figure~\ref{fig:prompt5}.

\subsection{Statistics of Filtered Dataset}
\label{sec:appendix_Statistics on Filtered questions}
\begin{figure}[]
    \centering
    \includegraphics[width=1\linewidth]{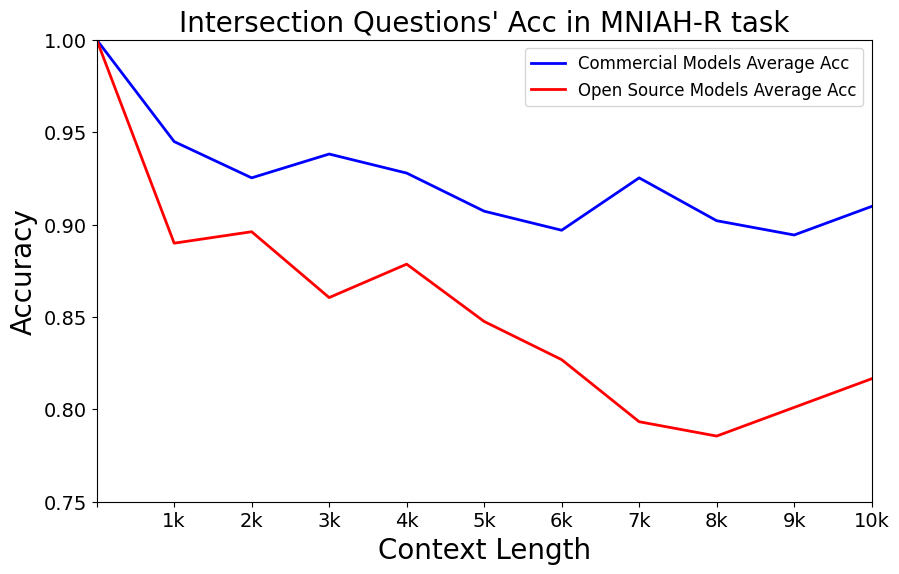}
    \caption{Performance of commercial models and open-source models on the intersection of filtered dataset.}
    \label{fig:Intersection}
\end{figure}

As illustrated in the Table~\ref{tab:statistic_of_filtered_question}, the total number of questions before filtering is 1582. After filtering, the number of questions varies slightly across models, though the overall distribution remains comparable. Specifically, GPT-4o contains the fewest questions, totaling 594, while Qwen2.5-7B-Instruct-1M has the most, with 886 questions. Furthermore, we calculate the intersection of questions across models after filtering, which amounts to 129, suggesting that the difficulty level of individual questions differs across models. To examine the impact of context length extension on model accuracy, it is essential to filter the questions for each model independently. We also calculated the accuracy on the intersection, and as shown in the Figure~\ref{fig:Intersection}, the performance of both the open-source and commercial models also decreases with increasing context length.

\begin{table}[]\footnotesize
\addtolength\tabcolsep{-1.0pt}
\begin{tabular}{lc|>{\columncolor{gray!20}}r}
\toprule
Models    & \texttt{\# After} &  \\ 
\cline{1-2}
GPT-4o    & 594   & \texttt{\# Before:} \\
GPT-4o-mini          & 637 & 1582 \\
Claude 3.5 Sonnet    & 776  &  \\
Llama-3-8B-ProLong   & 784 & \texttt{\# Intersection:} \\
Qwen2.5-7B-Instruct-1M          & 886 & 129 \\
GLM-4-9B-Chat-1M     & 884   &  \\ 
\bottomrule
\end{tabular}
\caption{Statistics of Filtered Questions. \texttt{\#After} denotes the number of questions remaining after filtering, \texttt{\#Before} represents the total number of questions before filtering, and \texttt{\#Intersection} refers to the number of overlapping questions across models after filtering.}
\label{tab:statistic_of_filtered_question}
\end{table}

\section{Training Details for Mitigation}
\subsection{Construction of Training and Test Dataset}
\label{sec:appendix_Construction_details_of_fine-tuning_dataset}
In Section~\ref{sec:Multi-Round Retrieval and Reasoning}, we analyze the test-time scaling law on the MNIAH-R task and find that performance generally saturates after the 2nd or 3rd round. Based on this, we select the first two rounds of reasoning from GPT-4o, which show the smallest accuracy decline, to construct the fine-tuning dataset. GPT-4o’s filtered questions consist of 594 items, from which we randomly sample 416 for the \texttt{training} set and 178 for the \texttt{test} set.

For each training question, we extract four responses: the initial retrieved information, the reasoned answer based on this information, the second retrieved answer after further reflection, and the reasoned answer derived from both retrievals. These responses are then submitted to GPT-4 for integration and rewriting, resulting in a cohesive thought process. The related prompt is illustrated in Figure~\ref{fig:prompt6}. During this step, the maximum generation length for GPT-4 is set to 512 tokens with a sampling temperature of 1. Three integrated results are generated for each question, yielding a total of 1,248 fluent thinking process entries. Following the method outlined in Appendix~\ref{sec:cons_MNIAH-R}, each question is subsequently extended to a length of 4,096 tokens, using the \texttt{Llama-3-8B-ProLong-64k-Instruct} encoder.

\subsection{Training Setting}
\label{sec:appendix_Training Setting Details}
The batch size for model training is set to 2, with a learning rate of 1e-5 and a warm-up rate of 0.03 for the cosine scheduler. The constructed dataset is trained for two epochs.

\section{Generation Settings of All Experiments}
\label{sec:appendix_Generation Settings of All Experiments}
We conduct multiple inference experiments, with some sharing the same generation parameters and others differing. To facilitate the community's review of our work, we provide a consolidated summary of all generation settings used in our experiments:
\begin{itemize}
    \item In the MNIAH-R task, for both filtered and unfiltered questions tests, as well as exploring the impact of the placement of needles and the distance between them on accuracy degradation tests, since the required length for answering multi-hop questions is relatively short, the model’s maximum generation length is set to 128 tokens, and using greedy decoding.
    \item In experiments investigating the variation of the model's thinking process length with increasing context, exploring the test-time scaling law of MNIAH-R, and tesing models fine-tuned with the thinking process, the maximum generation length is set to 512 tokens to ensure that the model's thinking process is not truncated, and using greedy decoding. In contrast, when testing models fine-tuned with direct answers, since the model does not need to output the thinking process, the maximum generation length is set to 128 tokens.
    \item In the mathematical application, when using GPT-4o to generate solutions, we set the sampling \texttt{temperature} to 1, \texttt{top\_p} to 0.95, and generate five responses per question. To avoid truncating the solutions, the model’s maximum generation length is set to 2048 tokens.
    \item When providing GPT-4o's solutions to the model fine-tuned with thinking process for testing, to avoid high output repetition~\citep{guo2025deepseek}, the sampling \texttt{temperature} is set to 0.6 and the \texttt{top\_p} value to 0.95, generating five responses per query, with a maximum generation length of 2048 tokens.
    \item DeepSeek-V3, as the evaluator, only requires evaluating correctness. Therefore, the model’s maximum generation length is set to 128 tokens, using greedy decoding.
\end{itemize}
All experiments are conducted with batch size set to 1 to avoid the impact of pad token on model performance.

\begin{figure*}[!h]
\begin{AIbox}{Prompt}
As an evaluator, your task is to evaluate \textbf{Model's Answer} according to the given \textbf{Reasoning Question} and \textbf{Correct Answer}.\\
\#\#\# Reasoning Question:\\
\texttt{Multi-Hop question}\\ \\
\#\#\# Model's Answer:\\
\texttt{Model's answer}\\ \\
\#\#\# Correct Answer:\\
\texttt{Ground Truth}\\ \\
\#\#\# Assessment Tasks\\
Determine whether the \textbf{Model's Answer} is correct based on the \textbf{Reasoning Question} and \textbf{Correct Answer}, return \textbf{1} if it is correct and \textbf{0} if it is incorrect.\\ \\
\#\#\# Answer Format:\\
Please answer in the following format: Assessment result: 0 or 1\\
\end{AIbox}
\caption{Prompt template for evaluating correctness of model's answer.}
\label{fig:prompt5} 
\end{figure*}

\section{Prompts of Experiments}
\label{sec:appendix_Prompts}

\begin{figure*}[!h]
\begin{AIbox}{Prompt}
<Question>: Multi-Hop question\\
\#\#\# Instruction\\
```\\
1. Based on the known information provided by <Question>, use tree-like thinking to reason step by step. Each step of reasoning should have a detailed problem-solving process and clear intermediate step calculation results.\\
2. You must give a final answer and put your final answer within /\/boxed{}.\\
```\\
\end{AIbox}
\caption{Prompt Template for generating a solution for the mathematical problem.}
\label{fig:prompt4} 
\end{figure*}

In the various experiments conducted in this paper, the prompts used for each experiment may differ. Therefore, in this section, we provide a summary of the prompts employed:
\begin{itemize}
    \item In the experiments that test the impact of filtering questions before and after on the MNIAH-R task, the investigation of how needle placement and the distance between needles affect accuracy degradation, the experiments comparing models fine-tuned with thinking processes versus those fine-tuned directly with answers, and the tests where GPT-4 solutions are provided to the fine-tuned model, the model only needs to answer the question based solely on the given context when generating responses. The prompt is shown in Figure~\ref{fig:prompt1}.
    \item In the experiment exploring the effect of context length on the model’s thinking process, we instruct the model to first provide a step-by-step reasoning process before presenting the final answer. The corresponding prompt is shown in Figure~\ref{fig:prompt2}.
    \item In the experiment exploring the test-time scaling law for the MNIAH-R task, we require the model to first retrieve useful information, then perform reasoning based on the retrieved message. For subsequent retrievals, reflection is introduced, and reasoning is conducted using all previously retrieved information, enabling iterative thinking. The prompt is shown in Figure~\ref{fig:prompt3}.
    \item In the application to mathematical scenarios using GPT-4o to generate problem-solving solutions, we require the model to adopt a tree-like thinking approach, providing detailed reasoning processes along with intermediate calculation results. The prompt is shown in Figure~\ref{fig:prompt4}.
    \item When evaluating the correctness of results with DeepSeek-V3, we combine the question statement with the standard answer to ensure a more accurate assessment. The prompt is shown in Figure~\ref{fig:prompt5}.
\end{itemize}
It is important to emphasize that the prompts provided in the figures represent only the "user" query content. The complete prompts should be constructed according to the respective chat template~\footnote{https://huggingface.co/docs/transformers/chat\_templating} of each model.

\begin{figure*}[]
\begin{AIbox}{Prompt}
\#\#\# Context\\
```\\
\{Multiple Needles in a Haystack\}\\
```\\
\#\#\# Instruction\\
```\\
Answer the Question based only on the information provided in the Context.\\
```\\
\#\#\# Question\\
```\\
Multi-Hop question\\
```
\end{AIbox}
\caption{Prompt Template for asking model to answer the question based solely on the context.}
\label{fig:prompt1} 
\end{figure*}

\begin{figure*}[]
\begin{AIbox}{Prompt}
\#\#\# Context\\
```\\
\{Multiple Needles in a Haystack\}\\
```\\
\\
<Question>: Multi-Hop question\\
\#\#\# Instruction\\
```\\
1. Answer only based on the information provided in the Context.\\
2. Please reason step by step, give your thought process and the answer to the <Question>.\\
3. Please answer in the following format:\\
Thought Process: <Step-by-step thinking process>\\
Answer: <The Answer to the Question>\\
```\\
\end{AIbox}
\caption{Prompt Template for asking the model to first provide a step-by-step reasoning process before presenting the final answer.}
\label{fig:prompt2} 
\end{figure*}

\begin{figure*}[]
\begin{AIbox}{Prompt of the Iteration of the thinking process}
\textbf{Prompt of First Retrive:}\\
\#\#\# Context\\
```\\
\{Multiple Needles in a Haystack\}\\
```\\
<Question>: Multi-Hop question\\
\#\#\# Instruction\\
1. Please accurately retrieve the information needed to answer the <Question> in the context as much as possible, and list them in points, with no less than 3 items. Just retrieve the information and do not answer the <Question>.\\
2. Please answer in the following format:\\
Evidence: <Retrieved Information>\\
\\
\textbf{Prompt of First Reason:}\\
<All Retrived Information>: First Retrived Information\\
<Question>: Multi-Hop question\\
\#\#\# Instruction\\
1. Please answer the <Question> based on the <All Retrived Information>.\\
2. Please answer in the following format:\\
Answer: <The answer to the question>\\
\\
\textbf{Prompt of Reflection and Retrive again:}\\
\#\#\# Context\\
```\\
\{Multiple Needles in a Haystack\}\\
```\\
<Question>: Multi-Hop question\\
<Last Time's Retrieved Information>: All the retrieved information concatenated together.\\
<Last Time's Answer>: The Last reasoning answer based on the retrieved information\\
\#\#\# Instruction\\
```\\
1. Your previous responses may be wrong. Now, please reflect on your previous responses and retrieve the information needed to answer <Question> from the Context again, while ensuring that it does not repeat information already in <Last Time's Retrieved Information>. List the information you retrieved this time, at least 3 items.\\
2. Just retrieve the information and do not answer the <Question>.\\
3. Please answer in the following format:\\
Evidence: <The Information Retrieved this Time>\\
```\\
\\
\textbf{Prompt of Reasoning based on all previously retrieved information:}\\
<All Retrived Information>: All the previous retrieved information concatenated together.\\
<Question>: Multi-Hop question\\
\#\#\# Instruction\\
1. Please answer the <Question> based on the <All Retrived Information>.\\
2. Please answer in the following format:\\
Answer: <The answer to the question>\\
\end{AIbox}
\caption{Prompt Template for exploring the test-time scaling law of the MNIAH-R task.}
\label{fig:prompt3} 
\end{figure*}

\begin{figure*}[]
\begin{AIbox}{Prompt}
\#\#\# Background\\
Now I am testing LLM on a multi-hop question answering task, where the evidence needed to answer the question is scattered in a context full of irrelevant information. \\
\\
\#\#\# The "4R" Method\\
My testing method follows the System 2 paradigm, which I call "4R"—"Retrieve, Reason, Retrieve again, Reason":\\
- 1R (Retrieve): The model first retrieves the information necessary to answer the Question from the Context.\\
- 2R (Reason): The model then answers the Question based on the information retrieved in 1R.\\
- 3R (Retrieve again): The model reflects on whether the results from 1R and 2R are correct, then retrieves the required information again, ensuring it does not repeat what was already retrieved in 1R.\\
- 4R (Reason): Finally, the model answers the Question based on the information from both 1R and 3R.\\
\\
\#\#\# Natural Thinking Process Generation\\
<Thought Process>\\
\{thought\_process\}\\
</Thought Process>\\
<Question>\\
\{question\}\\
</Question>\\
\\
The <Thought Process> above reflects the model’s reasoning based on the <Question> using the "4R" Method. Your task is to rewrite the <Thought Process> to resemble a more human-like, intuitive natural thinking process. The new version should:\\
1. Be presented as step-by-step reasoning:\\
\hspace*{0.5cm}1. The reasoning process including the preliminary retrieval in the context, initial answering, reflection, further retrieval in the context, and final answering.\\
\hspace*{0.5cm}2. The preliminary retrieval and further retrieval process must list all the retrieval information items in "1R (Retrieve)" and "3R (Retrieve again)" by points, like "1. 2. 3. ...".\\
2. Phrases that cannot be used: \\
\hspace*{0.5cm}1. Don't use the phrase "I remember". We are now answering the <Question> entirely based on context. \\
\hspace*{0.5cm}2. Don't use the terms like "1R (Retrieve)", "2R (Reason)", "3R (Retrieve again)", "4R (Reason)" and "step1", "step2", "step3", etc.\\
3. Focusing on natural transitions. Use casual and natural language for transitions, such as "hmm," "oh," "also," or "wait."\\
\\
Return directly the revised natural thinking in the following format:\\
“‘json\\
\{\{\\
"NaturalReasoning": "..."\\
\}\}\\
\end{AIbox}
\caption{Prompt Template for rewriting the iterative thinking process.}
\label{fig:prompt6}
\end{figure*}

\end{document}